\documentclass{article}
\usepackage{ijcai19}

\usepackage{times}
\usepackage{soul}
\usepackage{url}
\usepackage[hidelinks]{hyperref}
\usepackage[utf8]{inputenc}
\usepackage[small]{caption}
\usepackage{graphicx}
\usepackage{amsmath}
\usepackage{booktabs}
\usepackage{algorithm}
\usepackage{algorithmic}
\usepackage{epsfig}
\usepackage{amssymb}
\usepackage{mathtools}
\usepackage{multirow}
\usepackage{subcaption}
\usepackage{xcolor}
\usepackage{xspace}

\newcommand{\eat}[1]{}

\newcommand{\norm}[1]{\left\lVert#1\right\rVert}

\DeclareMathOperator*{\argmin}{arg\,min}

\makeatletter
\DeclareRobustCommand\onedot{\futurelet\@let@token\@onedot}
\def\@onedot{\ifx\@let@token.\else.\null\fi\xspace}

\def\eg{\emph{e.g}\onedot} 
\def\ie{\emph{i.e}\onedot} 
 
\def\etc{\emph{etc}\onedot} 
\def\wrt{w.r.t\onedot} 

\makeatother

\title{Deep Recurrent Quantization for Generating Sequential Binary Codes}

\author{
Jingkuan Song$^{1}$\and
Xiaosu Zhu$^{1}$\and
Lianli Gao$^{1}$\and
Xin-Shun Xu$^2$\and
Wu Liu$^3$\And
Heng Tao Shen$^{1}$\footnote{Contact Author}\\
\affiliations
$^1$Center for Future Media, University of Electronic Science and Technology of China\\
$^2$Shandong University\\
$^3$JD AI Research\\
\emails
jingkuan.song@gmail.com,
xiaosu.zhu@outlook.com,
lianli.gao@uestc.edu.cn,
xuxinshun@sdu.edu.cn,
liuwu1@jd.com,
shenhengtao@hotmail.com
}

\begin{document}

\maketitle

\begin{abstract}
Quantization has been an effective technology in ANN (approximate nearest neighbour) search due to its high accuracy and fast search speed. To meet the requirement of different applications, there is always a trade-off between retrieval accuracy and speed, reflected by variable code lengths. However, to encode the dataset into different code lengths, existing methods need to train several models, where each model can only produce a specific code length.  This incurs a considerable training time cost, and largely reduces the flexibility of quantization methods to be deployed in real applications. To address this issue, we propose a Deep Recurrent Quantization (\textit{DRQ}) architecture which can generate sequential binary codes. To the end, when the model is trained, a sequence of binary codes can be generated and the code length can be easily controlled by adjusting the number of recurrent iterations. A shared codebook and a scalar factor is designed to be the learnable weights in the deep recurrent quantization block, and the whole framework can be trained in an end-to-end manner. As far as we know, this is the first quantization method that can be trained once and generate sequential binary codes. Experimental results on the benchmark datasets show that our model achieves comparable or even better performance compared with the state-of-the-art for image retrieval. But it requires significantly less number of parameters and training times. Our code is published online: \url{https://github.com/cfm-uestc/DRQ}.

\end{abstract}

\section{Introduction}

\begin{figure*}[t]
    \begin{center}
        \includegraphics[width=0.7\linewidth]{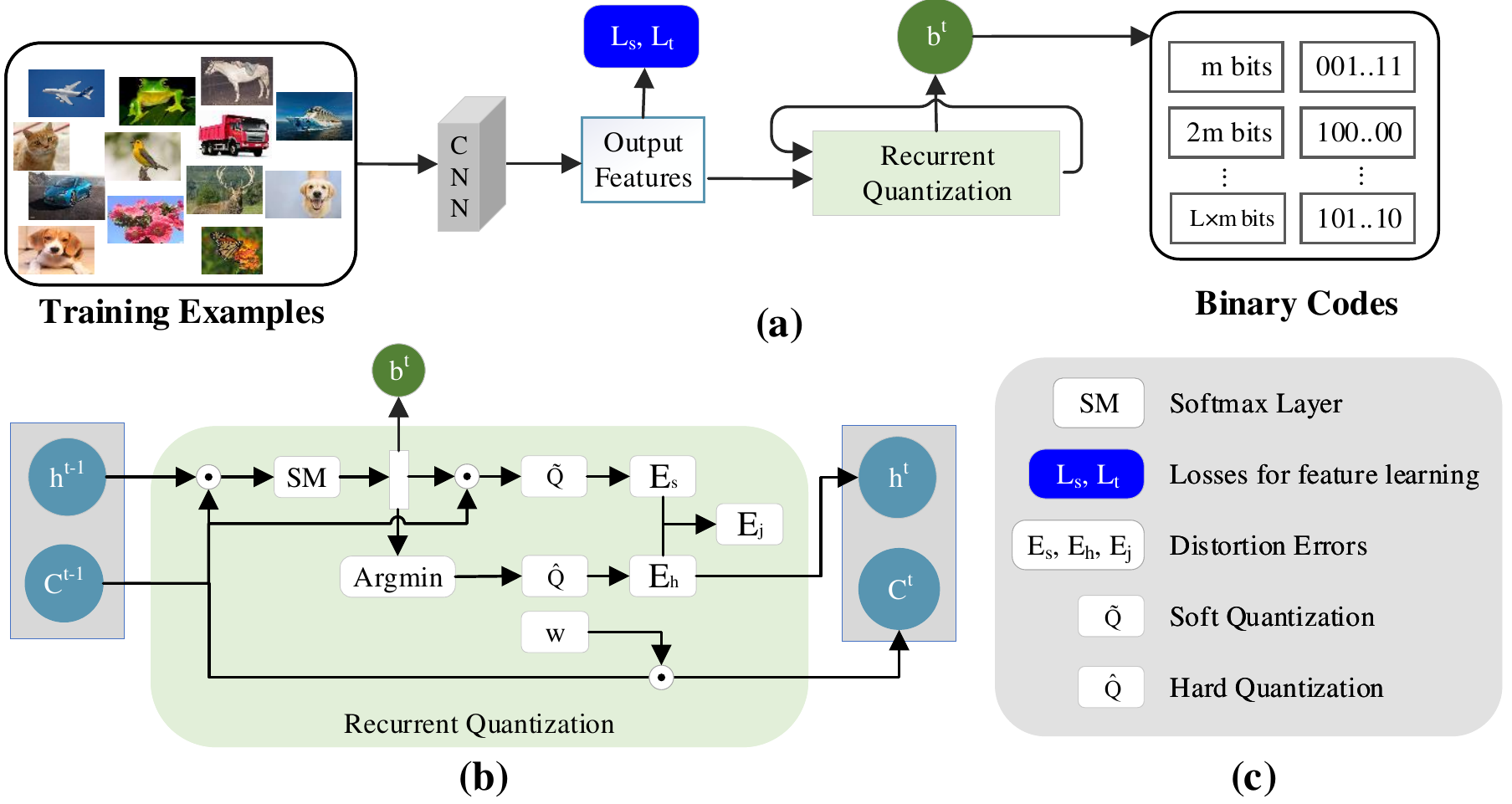}
    \end{center}
	\vspace{-0.4cm}
    \caption{Illustration of our Deep Recurrent Quantization network architecture. The whole framework is depicted in (a). It contains two main models, feature refinement and recurrent quantization block (b). DRQ can generate a sequence of binary codes.}
    \label{fig:whole_structure}
\end{figure*}

With the significant increase of the mass media contents, image retrieval has become the highly-concerned spot. Image retrieval concentrates on searching similar images from large-scale database. The direct way is to use reliable kNN (k-nearest neighbor) techniques, which usually perform brute-force searching on database. ANN (approximate nearest neighbor) search is an optimized algorithm which is actually practicable against kNN search. The main idea of ANN search is to find a compact representation of raw features \ie a binary code with fixed length, which can retain structure of raw feature space and dramatically improve the computation speed. 

Recently, hashing methods have been widely used in ANN search. They usually learn a hamming space which is refined to maintain similarity between features~\cite{DSH,DHN,VideoH,SongHGXHS18,QBH}. Since the computation of hamming distance is super fast, hashing methods have huge advantages on ANN search. However, hashing methods lack accuracy on feature restoration. Methods based on quantization require a codebook to store some representative features. Therefore, the main goal of quantization is to reserve more information of feature space in codebook. Then, they try to find a combination of codewords to approximate raw features and only to store indexes of these codewords.

Quantization is originated from $k$-means algorithm, which first clusters data points and uses the clustering centers as codebook. Each data point is represented by index of its corresponding center. In order to decrease computation cost of $k$-means, product quantization~\cite{ProductQuantization} and optimized product quantization~\cite{OptimizedProductQuantization} split whole feature space into a set of sub-regions and perform similar algorithm on each subspace respectively. Such initial quantization methods construct restrictions and well-designed codebooks to accelerate calculation. In the deep learning era, people proposed some end-to-end deep neural networks to perform image feature learning and quantization together. Deep quantization network~\cite{DQN} use AlexNet to learn well-separated image features and use OPQ to quantize features. Deep visual-semantic quantization~\cite{DVSQ} and deep triplet quantization~\cite{DTQ} quantize features by CQ. Different from these works, product quantization network~\cite{PQNet} proposed a differentiable method to represent quantization as operations of neural network, so that gradient descent can be applied to quantization. Despite their successes, PQ and its variants have several issues. First, to generate binary codes with different code lengths, a retraining is usually unavoidable. Second, it is tricky for the decomposition of high-dimensional vector space. Different decomposition strategies may result in huge performance differences. To tackle these issues, we propose a deep quantization method called deep recurrent quantization, which constructs codebook that can be used recurrently to generate sequential binary codes. Extensive experiments show our method outperforms state-of-the-art methods even though they use larger codebooks.

\section{Preliminaries}

Quantization-based image retrieval tasks are defined as follows: Given a set of images $\mathbb{I} \in {\{0, 1, \cdots, 255\}}^{N \times H \times W \times C}$ which contain $N$ images of height $H$, width $W$ and channel $C$. We first use a CNN, \eg AlexNet and VGG to learn a hyper representation $\textbf{X} \in \mathbb{R}^{N \times D}$ of images, where $D$ is the dimension of feature vectors. Then we apply quantization on these feature vectors, to learn a codebook $\textbf{C} \in \mathbb{R}^{K \times D}$ which contains $K$ codewords and each of them has $D$ dimensions. Feature vectors are then compressed to compact binary codes $\textbf{B} \in {\{0, 1\}}^{N \times L}$ where $L$ indicates the code length.

\subsection{Integrate Quantization To Deep Learning Architectures}
\label{sec:IntegratedQuantization}
During the procedure of quantization, to pick a closest codeword from feature representation is to compute the distance between codewords and features and find the minimum one, which can be described as:
\begin{equation}
    \label{eq:1}
    \hat{\textbf{q}} = \hat{\textbf{Q}}(\textbf{x}) = \argmin_{\textbf{C}_k}{\norm{\textbf{C}_k - \textbf{x}}_2}, i=1,2,\cdots,K
\end{equation}
where $\textbf{x}$ are the features of a data point, and $\textbf{C}_k$ is the $k$-th codeword, $\hat{\textbf{Q}}(\textbf{x})$ is quantization function and $\hat{\textbf{q}}$ is quantized feature.
Therefore, $\hat{\textbf{q}}$ is the approximation of $\textbf{x}$. Meanwhile, we collect the index of codeword as the quantized code, which is described as:
\begin{equation}
    \label{eq:2}
    \textbf{b} = \argmin_{k}{\norm{\textbf{C}_k - \textbf{x}}_2}, k=1,2,\cdots,K
\end{equation}
Since $\textbf{b}$ is in the range of $0 \sim$ $K$-$1$, then all the codes can be binarized to a code length of $\log_2{K}$. Then, the original feature $\textbf{x}$ can be compressed to an extremely short binary code.

However, the formulation of codeword is non-differentiable, \ie $\frac{\partial \hat{\textbf{q}}}{\partial \textbf{C}}$ does not exist. It cannot be directed integrated into deep learning architectures. To tackle this issue, we use a convex combination of codewords to approximate features, which is defined as follows:
\begin{align}
    &p_k(\textbf{x}) = \frac{e^{-\gamma \norm{\textbf{C}_k - \textbf{x}}_2}}{\sum_{j=1}^{K}{e^{-\gamma \norm{\textbf{C}_j - \textbf{x}}_2}}}\\
    &\tilde{\textbf{q}} = \tilde{\textbf{Q}}(\textbf{x}) = \sum\nolimits_{k=1}^{K}{p_k(\textbf{x}) \textbf{C}_k}
\end{align}
Here, $p_k(\textbf{x})$ indicates the confidences of each codewords \wrt~{$\textbf{x}$}, \ie the closer one codeword $\textbf{C}_k$ is to a feature $\textbf{x}$, the higher $p_k(\textbf{x})$ will be. Then, $\hat{\textbf{q}}$ is approximated by $\tilde{\textbf{q}}$, which is the weighted sum of all codewords. We define $\hat{\textbf{Q}}(\textbf{x})$ as hard quantization and $\tilde{\textbf{Q}}(\textbf{x})$ as soft quantization.

\section{Proposed Method} 

The whole network architecture of our deep recurrent quantization (DRQ) is demonstrated in Fig.~\ref{fig:whole_structure}. DRQ contains two main parts: feature extraction module and quantization module. In feature extraction module, we apply intermediate supervision on top of CNN, to guide the learning of semantic-embedded visual features. In quantization module, we design a recurrent quantization block and integrate it into deep learning architecture which can be trained end-to-end.

\subsection{Intermediate Supervision for Features}
To get the feature representation of images, we use AlexNet to extract features from the last linear layer. To leverage the clustering performance \ie, to let the images with the same label have higher similarity and vice versa, we apply two losses with intermediate supervision. Specifically, we first collect a triplet in dataset which contains an anchor image $I^o$, a positive sample $I^+$ and a negative sample $I^-$ \wrt{~anchor} (for multi-label images, we define a positive image as one which shares at least one label with anchor, and a negative image as one which does not share any label with an anchor), and feed them into AlexNet to obtain the 4096-d features from $fc7$ layer. Then we add two linear layers $fc8$ of 1748-d and $fc9$ of 300-d. We concatenate $fc8$ and $fc9$ to get a final feature $\textbf{x}$ of 2048-d. Since we feed the triplet into the network, the output features are represented as $\textbf{x}^o, \textbf{x}^+, \textbf{x}^-$.

We apply two supervised objective function on these layers: 1) Adaptive margin loss $\ell_s$, which is from DVSQ~\cite{DVSQ} and applied to $fc9$ outputs of triplet, and 2) Triplet loss $\ell_t$ defined to final feature $\textbf{x}$, which is a concatenated feature of $fc8$ and $fc9$. $\ell_s$ is defined as:
\begin{align}
    \ell_s(\textbf{x})\! &= \!\sum_{i \in \mathcal{Y}_n} \sum_{j \notin \mathcal{Y}_n} (0, \delta_{ij} \!-\! \frac{\boldsymbol{v_i^\intercal}\boldsymbol{z}_n}{\norm{\boldsymbol{v}_i}\norm{\boldsymbol{z}_n}} \!+\! \frac{\boldsymbol{v_j^\intercal}\boldsymbol{z}_n}{\norm{\boldsymbol{v}_j}\norm{\boldsymbol{z}_n}}) \nonumber\\
    \delta_{ij} &= 1 - \frac{\boldsymbol{v}_i^\intercal \boldsymbol{v}_j}{\norm{\boldsymbol{v}_i}\norm{\boldsymbol{v}_j}}
\end{align}

Triplet loss $\ell_t$ can adjust features to adapt to clustering, which uses the triplet of $x^o, x^+, x^-$. It is defined as:
\begin{equation}
    \ell_t(\textbf{x}^o\!, \textbf{x}^+\!, \textbf{x}^-) \!=\! \max{(\norm{\textbf{x}^o \!-\! \textbf{x}^+}_2\! -\! \norm{\textbf{x}^o \!-\! \textbf{x}^-}_2 \!+\!\delta, 0)}
\end{equation}

\subsection{Recurrent Quantization Block}

In recurrent quantization model, we adopt a shared codebook that contains $K$ codewords. We denote the level of quantization code as $M$, which indicates how many iterations the codebook is reused. For each level, we pick a proper codeword as the approximation of feature vectors, and we take the index of picked codeword as the quantization code. For example, if we set $K=256, M=4$, the index range of each level quantization code is $0-255$, represented as a binary code of $\log_2{K}$=$8$ bits. The total length of quantization code is $M \times \log_2{K} = 4 \times 8 = 32$. The position $0-7$ is the index of first level codeword, $8-15$ is the index of second level, \etc. Therefore, the feature vector can be approximated by a combination of a few codewords in the codebook.

As we described in Sec.~\ref{sec:IntegratedQuantization}, to perform a quantization, input $\textbf{x}$ and codebook $\textbf{C}$ are necessary. Output is the code $\textbf{b}$. Inspired by the hierarchical codebooks in stacked quantizer~\cite{StackedQuantizer}, we observe the residual of $\textbf{x}$ can be used as an input to the next quantizer. Therefore, a basic idea is to perform quantization step-by-step:
\begin{align}
    \hat{\textbf{q}}^1 &= \hat{\textbf{Q}}^1(\textbf{x}) \nonumber,~
    \textbf{r}^1 = \textbf{x} - \hat{\textbf{q}}^1 \nonumber \\
    \hat{\textbf{q}}^2 &= \hat{\textbf{Q}}^2(\textbf{r}^1) \nonumber,~
    \textbf{r}^2 = \textbf{x} - \hat{\textbf{q}}^1 - \hat{\textbf{q}}^2 \nonumber \\
    &\;\;\vdots \nonumber \\
    \hat{\textbf{q}}^M &= \hat{\textbf{Q}}^M(\textbf{r}^{M-1})
    \label{eq:stacked}
\end{align}
Specifically, $\hat{\textbf{q}}^m$ is quantized feature explained above, and $\textbf{r}^1, \textbf{r}^2, \cdots, \textbf{r}^{M-1}$ are residuals of $\textbf{x}$\eat{, \ie $\textbf{x} - \sum_i{\hat{\textbf{q}}^i}$}. We put $\textbf{r}^m$ to the next quantization to get the $\textbf{q}^{m+1}$ which approximates $\textbf{r}^m$. Therefore, $\sum_m{\textbf{q}^m}$ can be described as an approximation of $\textbf{x}$, which is much preciser than $\hat{q}^1$. Notice that processing of $\textbf{q}^m$ is similar. If we use a shared codebook, the computation in Eq.~\ref{eq:stacked} can be rewritten recurrently:
\begin{align}
    &\textbf{b}^m = \argmin_{k}{\norm{\textbf{C}_k - \textbf{h}^{m-1}}_2}, k=1,2,\cdots,K, m \geq 0 \nonumber \\
    &\textbf{h}^m = \textbf{h}^{m-1} - \textbf{C}^{m-1}_{\textbf{b}_{m-1}}, 1 \leq m \leq M       \nonumber \\
    &\textbf{C}^m = {w} \times \textbf{C}^{m-1}, 1 \leq m \leq M      \nonumber \\
    &\textbf{h}^0 = \textbf{x}, \textbf{C}^0=\textbf{C},
\end{align}
And the soft and hard quantization of $\textbf{h}^{m-1}$ is defined as:
\begin{align}
    \hat{\textbf{q}}^m &= \hat{\textbf{Q}}(\textbf{h}^{m-1}), 1 \leq m \leq M \nonumber \\
    \tilde{\textbf{q}}^m &= \tilde{\textbf{Q}}(\textbf{h}^{m-1}), 1 \leq m \leq M
\end{align}

\begin{figure}[t]
    \begin{center}
        \includegraphics[width=0.8\columnwidth]{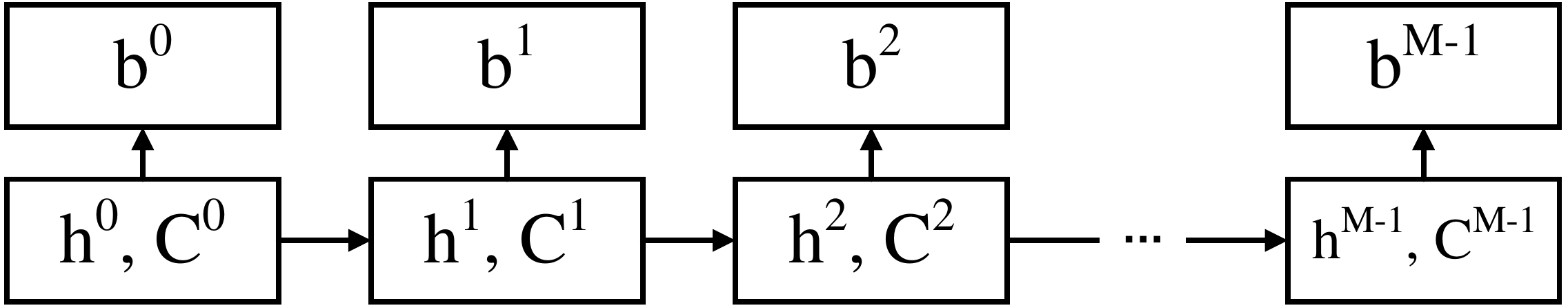}
    \end{center}
	\vspace{-0.3cm}
    \caption{The unfolded version of recurrent quantization.}
    \label{fig:recurrent_quantization}
\end{figure}

The unfolded structure of recurrent quantization is depicted in Fig.~\ref{fig:recurrent_quantization}. Here, $\textbf{h}^0, \textbf{C}^0$ is the raw features and initial codebook. $w$ is a shared learnable parameter with random initialization. In iteration $m$, we compute $\textbf{b}^m$ to find the best-fitted codeword, then we use $\textbf{C}^m, \textbf{b}^m, \textbf{h}^m$ to compute residual of $\textbf{h}^m$ and treat residual as next input $\textbf{h}^{m+1}$. Since the residual is one or more order of magnitudes lower than $\textbf{h}^m$, the next input should be much smaller than codewords in codebook, so we use $w \in \mathbb{R}$ as a scale factor to adjust the norm of codebook in order to fit the new input. In next iteration $m+1$, we use the scaled codebook $\textbf{C}^{m+1}$ to complete another similar computation. Finally, we learn a codebook $\textbf{C}$, a scale factor $w$ and sequential binary codes $\textbf{b}^0, \textbf{b}^1, \cdots, \textbf{b}^M$. The hard and soft quantization of $\textbf{x}$ can be computed as:
\begin{align}
    \hat{\textbf{x}} &= \hat{\textbf{q}}^0 + \hat{\textbf{q}}^1 + \cdots + \hat{\textbf{q}}^M       \nonumber \\
            &= \textbf{C}^0_{\textbf{b}_0} + w \times \textbf{C}^1_{\textbf{b}_1} + \cdots + w^{M-1} \times \textbf{C}^M_{\textbf{b}_M} \\
    \tilde{\textbf{x}} &= \tilde{\textbf{q}}^0 + \tilde{\textbf{q}}^1 + \cdots + \tilde{\textbf{q}}^M
\end{align}

By reusing codebook $\textbf{C}$, we can reduce the number of parameters by $M$ times.

\subsubsection{Objective Function}
Since $\hat{\textbf{q}}(\textbf{x})$ and $\tilde{\textbf{q}}(\textbf{x})$ are approximation of feature $\textbf{x}$, we define a distortion error as:
\begin{align}
    E^m_h\! = \!||\sum\nolimits_{i=1}^{m}{\hat{\textbf{q}}^m} \!-\! \textbf{x}||_2,~
    E^m_s\! = \!||\sum\nolimits_{i=1}^{m}{\tilde{\textbf{q}}^m} \!-\! \textbf{x}||_2
\end{align}
where $E^m_h$ is the distortion error between $\hat{\textbf{q}}^m$ and $\textbf{x}$ at iteration $m$ and $E^m_s$ is the distortion error between $\tilde{\textbf{q}}^m$ and $\textbf{x}$. We sum distortions for each level and the total distortion error is:
\begin{align}
    E_h = \sum\nolimits_{m=1}^M{E^m_h}, ~
    E_s = \sum\nolimits_{m=1}^M{E^m_s}
\end{align}

We also design a joint central error $E_j$ to align $E_h$ and~$E_s$:
\begin{equation}
    E_j = \norm{E_h - E_s}_2
\end{equation}

\subsection{Optimization}
In DRQ, there are two main losses: 1) $\ell_t$ and $\ell_s$ which refine features, 2) $E_h, E_s, E_j$, which control the quantization effectiveness. We split the training procedure into three stages. Firstly, we minimize $\ell_t, \ell_s$ together to pre-train our preceding neural network. Then, we add recurrent quantization block into network but only perform one recurrent iteration \ie set $M=1$ and optimize $\ell_t, \ell_s, E_h, E_s, E_j$ together. This is to get an initial codebook which are optimized for short binary codes. Finally, we set $M$ to a specified value and optimize the whole network with all losses, until it converges or we reach the max number of training iterations.

\section{Experiments}
To validate the effectiveness and efficiency of our adopted deep recurrent quantization, we perform extensive experiments on three public datasets: \textbf{CIFAR-10}, \textbf{NUS-WIDE} and \textbf{ImageNet}. \textit{Since existing methods use different settings, to make a thorough comparison with them, we follow these works and compare with them using separate settings.} 
We implement our model with Tensorflow, using a pre-trained AlexNet and construct intermediate layers on top of the $fc7$ layer. Meanwhile, we randomly initialize codebook with specified $M$ and $K$, which will be described below. We use Adam optimizer with $lr=0.001, \beta_1=0.9, \beta_2=0.999$ for training.

\subsection{Comparison Results Using Setting 1}
\subsubsection{Settings}
We first conduct results and make comparisons with state-of-the art methods on two benchmark datasets: \textbf{CIFAR-10} and \textbf{NUS-WIDE}. \textbf{CIFAR-10} is a public dataset labeled in 10 classes. It consists of 50,000 images for training and 10,000 images for validation. We follow~\cite{PQNet} to combine the training and validation set together, and randomly sample 5,000 images per class as database. The remaining 10,000 images are used as queries. Meanwhile, we use the whole database to train the network. \textbf{NUS-WIDE} is a public dataset consisting of 81 concepts, and each image is annotated with one or more concepts. We follow~\cite{PQNet} to use the subset of 195,834 images from the 21 most frequent concepts. We randomly sample 1,000 images per concept as the query set, and use the remaining images as the database. Furthermore, we randomly sample 5,000 images per concept from the database as the training set. We use mean Average Precision (mAP@5000) as the evaluation metric.

\begin{table}[t]
	\centering
	\small
	\begin{tabular}{c|cccc}
		\hline
		Method & 16 bits & 24 bits & 36 bits & 48 bits \\ \hline
		DRSCH  & 0.615   & 0.622   & 0.629   & 0.631   \\
		DSCH   & 0.609   & 0.613   & 0.617   & 0.686   \\
		DSRH   & 0.608   & 0.611   & 0.617   & 0.618   \\
		VDSH   & 0.845   & 0.848   & 0.844   & 0.845   \\
		DPSH   & 0.903   & 0.885   & 0.915   & 0.911   \\
		DTSH   & 0.915   & 0.923   & 0.925   & 0.926   \\
		DSDH   & 0.935   & 0.940   & 0.939   & 0.939   \\
		PQNet  & \textbf{0.947}   & \textbf{0.947}   & \textbf{0.946}   & \textbf{0.947}   \\ \hline
		DRQ    & \underline{0.942}   & \underline{0.943}   & \underline{0.943}   & \underline{0.943}   \\ \hline
	\end{tabular}
	\caption{Retrieval performance on CIFAR-10. The scores reported are mean Average Precision values.}
	\label{tab:eccv_cifar}
\end{table}

\begin{table}[t]
	\centering
	\small
	\begin{tabular}{c|cccc}
		\hline
		Method & 12 bits & 24 bits & 36 bits & 48 bits \\ \hline
		SH     & 0.621   & 0.616   & 0.615   & 0.612   \\
		ITQ    & 0.719   & 0.739   & 0.747   & 0.756   \\
		LFH    & 0.695   & 0.734   & 0.739   & 0.759   \\
		KSH    & 0.768   & 0.786   & 0.790   & 0.799   \\
		SDH    & 0.780   & 0.804   & 0.815   & 0.824   \\
		FASTH  & 0.779   & 0.807   & 0.816   & 0.825   \\ \hline
		NINH   & 0.674   & 0.697   & 0.713   & 0.715   \\
		DHN    & 0.708   & 0.735   & 0.748   & 0.758   \\
		DQN    & 0.768   & 0.776   & 0.783   & 0.792   \\
		DPSH   & 0.752   & 0.790   & 0.794   & 0.812   \\
		DTSH   & 0.773   & 0.808   & 0.812   & 0.824   \\
		DSDH   & \underline{0.776}   & 0.808   & 0.820   & 0.829   \\
		PQNet  & \textbf{0.795}   & \underline{0.819}   & \underline{0.823}   & \underline{0.830}   \\ \hline
		DRQ    & 0.772   & \textbf{0.838}   & \textbf{0.840}   & \textbf{0.843}   \\ \hline
	\end{tabular}
	\caption{Retrieval performance on NUS-WIDE. The scores reported are mean Average Precision values.}
	\label{tab:eccv_nus}
\end{table}

\begin{table*}[t]
	\small
	\centering
	\resizebox{0.95\textwidth}{!}{%
		\begin{tabular}{c|cccc|cccc|cccc}
			\hline
			\multirow{2}{*}{Method} & \multicolumn{4}{c|}{CIFAR-10} & \multicolumn{4}{c|}{NUS-WIDE} & \multicolumn{4}{c}{ImageNet} \\ \cline{2-13}
			& 8 bits & 16 bits & 24 bits & 32 bits & 8 bits & 16 bits & 24 bits & 32 bits & 8 bits  & 16 bit & 24 bits & 32 bits \\ \hline
			ITQ-CCA     & 0.315 & 0.354 & 0.371 & 0.414 & 0.526 & 0.575 & 0.572 & 0.594 & 0.189 & 0.270 & 0.339 & 0.436 \\
			BRE         & 0.306 & 0.370 & 0.428 & 0.438 & 0.550 & 0.607 & 0.605 & 0.608 & 0.251 & 0.363 & 0.404 & 0.453             \\
			KSH         & 0.489 & 0.524 & 0.534 & 0.558 & 0.618 & 0.651 & 0.672 & 0.682 & 0.228 & 0.398 & 0.499 & 0.547             \\
			SDH         & 0.356 & 0.461 & 0.496 & 0.520 & 0.645 & 0.688 & 0.704 & 0.711 & 0.385 & 0.516 & 0.570 & 0.605             \\
			SQ          & 0.567 & 0.583 & 0.602 & 0.615 & 0.653 & 0.691 & 0.698 & 0.716 & 0.465 & {0.536} & \underline{0.592} & \underline{0.611} \\ \hline
			CNNH        & 0.461 & 0.476 & 0.465 & 0.472 & 0.586 & 0.609 & 0.628 & 0.635 & 0.317 & 0.402 & 0.453 & 0.476             \\
			DNNH        & 0.525 & 0.559 & 0.566 & 0.558 & 0.638 & 0.652 & 0.667 & 0.687 & 0.347 & 0.416 & 0.497 & 0.525             \\
			DHN         & 0.512 & 0.568 & 0.594 & 0.603 & 0.668 & 0.702 & 0.713 & 0.716 & 0.358 & 0.426 & 0.531 & 0.556             \\
			DSH         & 0.592 & 0.625 & 0.651 & 0.659 & 0.653 & 0.688 & 0.695 & 0.699 & 0.332 & 0.398 & 0.487 & 0.537             \\
			DVSQ        & 0.715 & 0.727 & 0.730 & 0.733 & 0.780 & 0.790 & 0.792 & 0.797 & 0.500 & 0.502 & 0.505 & 0.518             \\
			DTQ         & \underline{0.785} & \underline{0.789} & \underline{0.790} & \underline{0.792} & \textbf{0.795} & \underline{0.798} & \underline{0.800} & \underline{0.801} & \textbf{0.641} & \textbf{0.644} & \textbf{0.647} & \textbf{0.651} \\ \hline
			DRQ         & \textbf{0.803}    & \textbf{0.824}    & \textbf{0.832}   & \textbf{0.837}    & \underline{0.748}    & \textbf{0.810}    & \textbf{0.817}    & \textbf{0.821}    & \underline{0.551} & \underline{0.583}    & {0.585}    & {0.587}    \\ \hline
		\end{tabular}
	}
	\caption{Quantitative comparison with state-of-the-art methods on three datasets. The scores reported are mean Average Precision values.}
	\label{tab:Result}
\end{table*}

\begin{figure*}[t]
	\begin{center}
		\includegraphics[width=0.82\paperwidth]{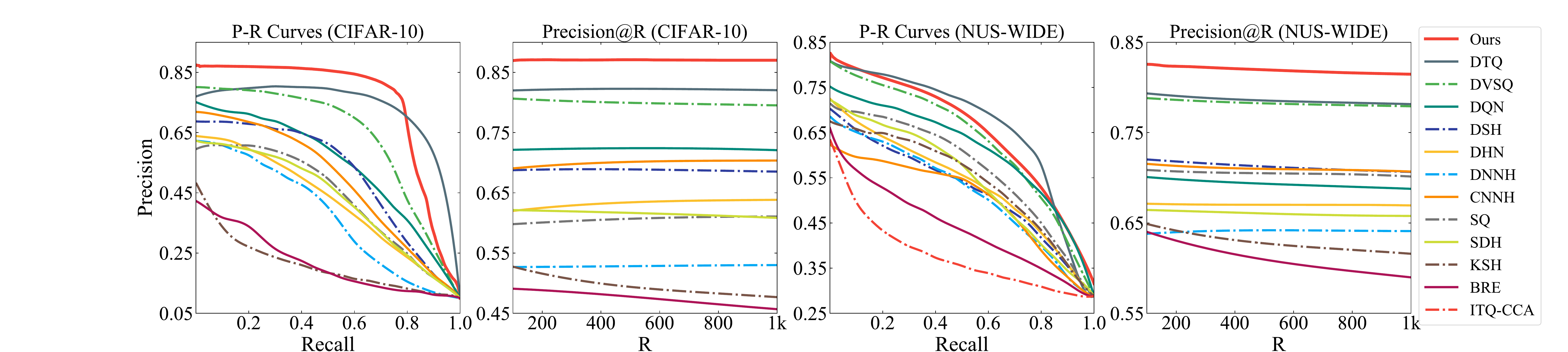}
		\vspace{-0.6cm}
		\caption{Quantitative comparison with state-of-the-art methods on two datasets: CIFAR-10 and NUS-WIDE. For each data set, we demonstrate Precision-Recall (P-R) curves and Precision@R curves. All results are based on 32-bit.}
		\label{fig:CurveOnFirst}
	\end{center}
\end{figure*}

\begin{figure}[t]
	\begin{center}
		\includegraphics[width=1\columnwidth]{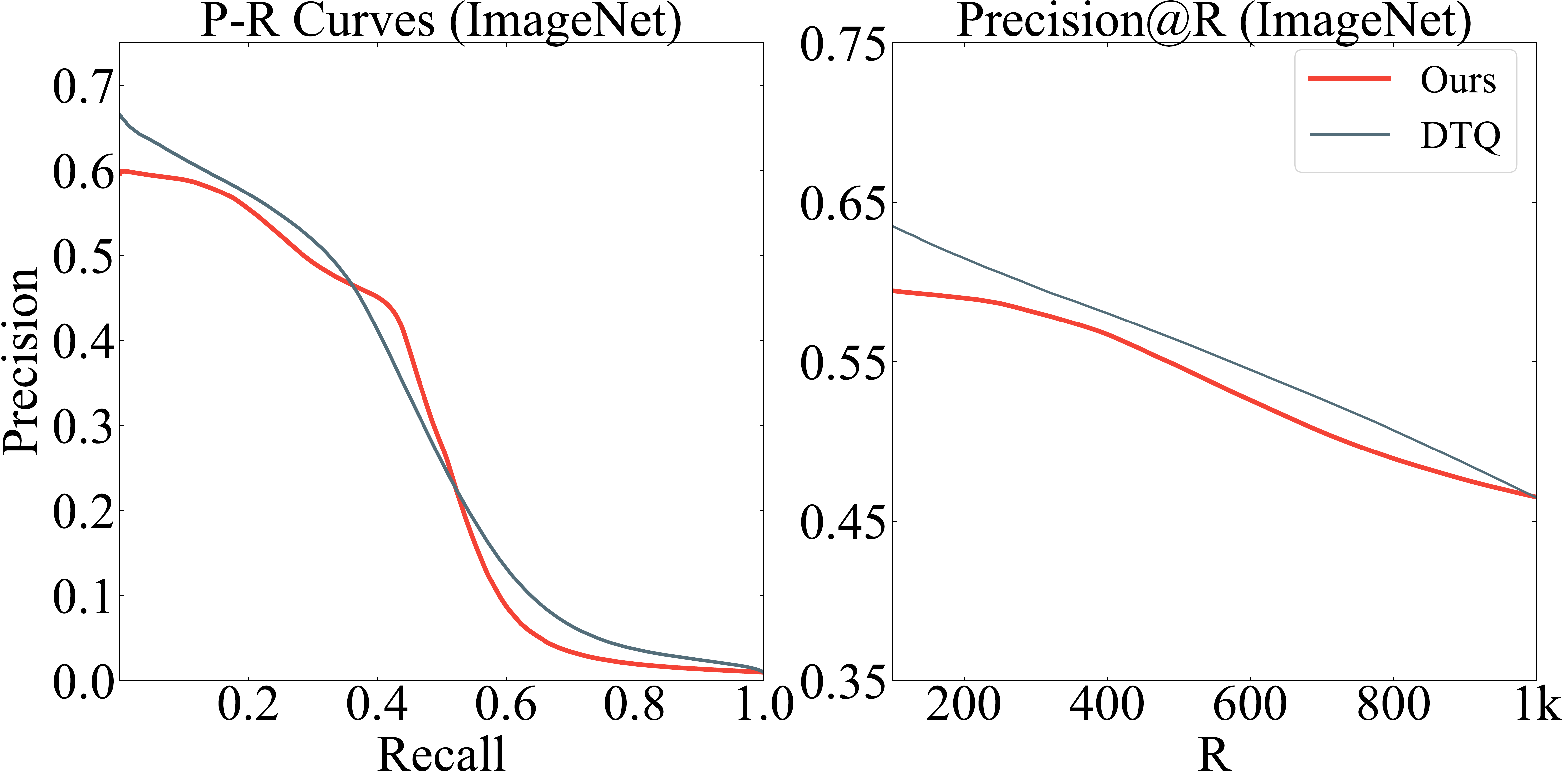}
		\vspace{-0.6cm}
		\caption{Comparison on ImageNet. All results are based on 32-bit.}
		\label{fig:CurveOnSecond}
	\end{center}
\end{figure}

\begin{figure*}[t]
	\begin{center}
		\begin{subfigure}{.18\paperwidth}
			\centering
			\includegraphics[width=1\linewidth,height=0.8\linewidth]{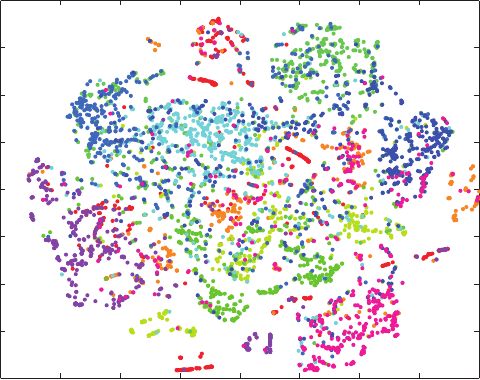}
			\caption{DVSQ}
		\end{subfigure}
		\begin{subfigure}{.18\paperwidth}
			\centering
			\includegraphics[width=1\linewidth,height=0.8\linewidth]{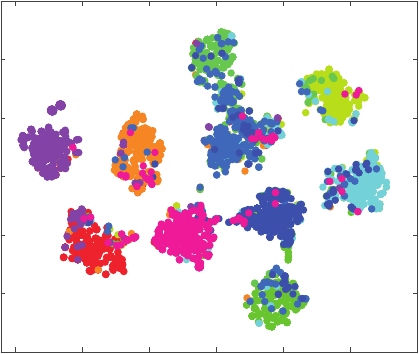}
			\caption{DTQ}
		\end{subfigure}
		\begin{subfigure}{.18\paperwidth}
			\centering
			\includegraphics[width=1\linewidth,height=0.8\linewidth]{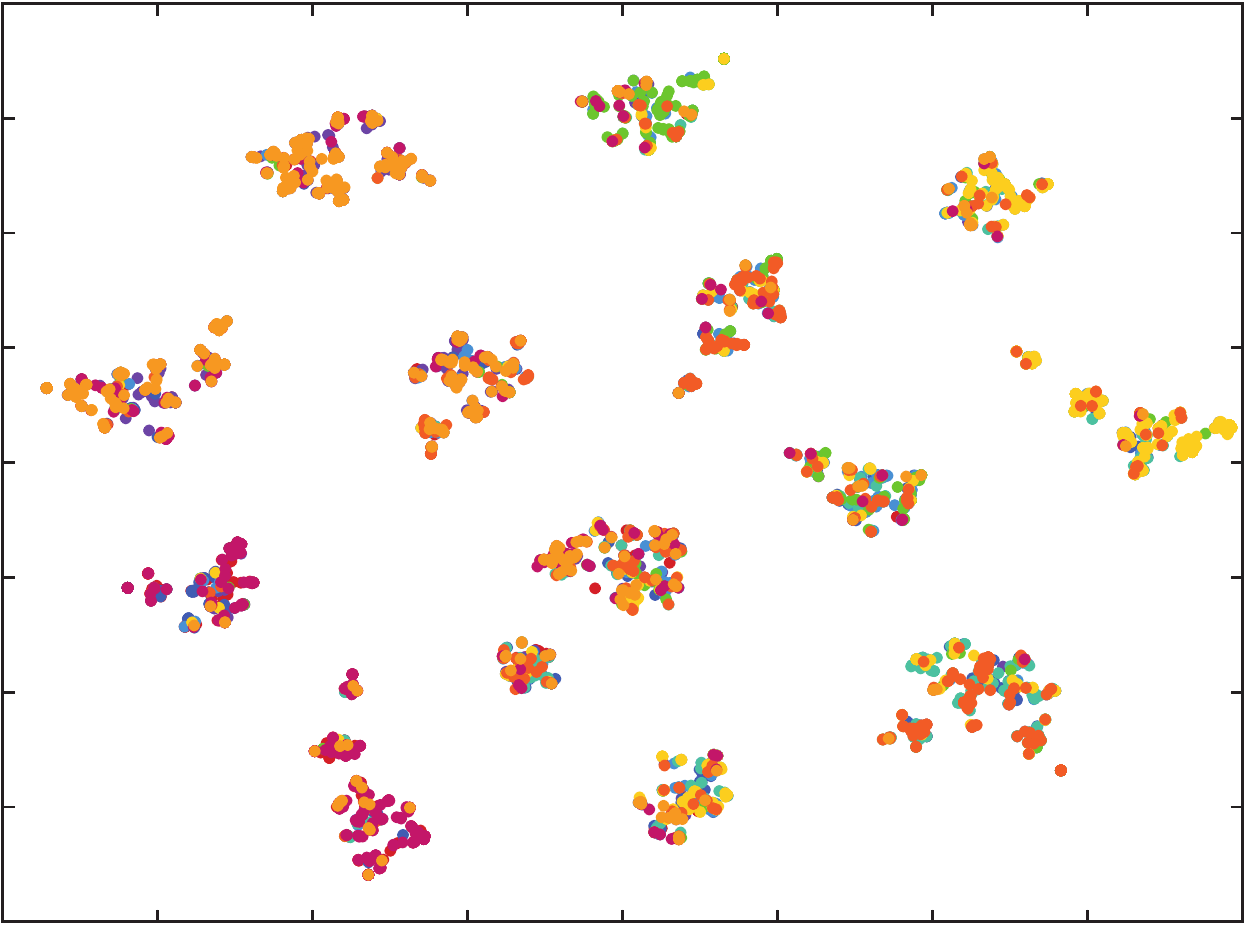}
			\caption{Our Unsupervised}
		\end{subfigure}
		\begin{subfigure}{.18\paperwidth}
			\centering
			\includegraphics[width=1\linewidth,height=0.8\linewidth]{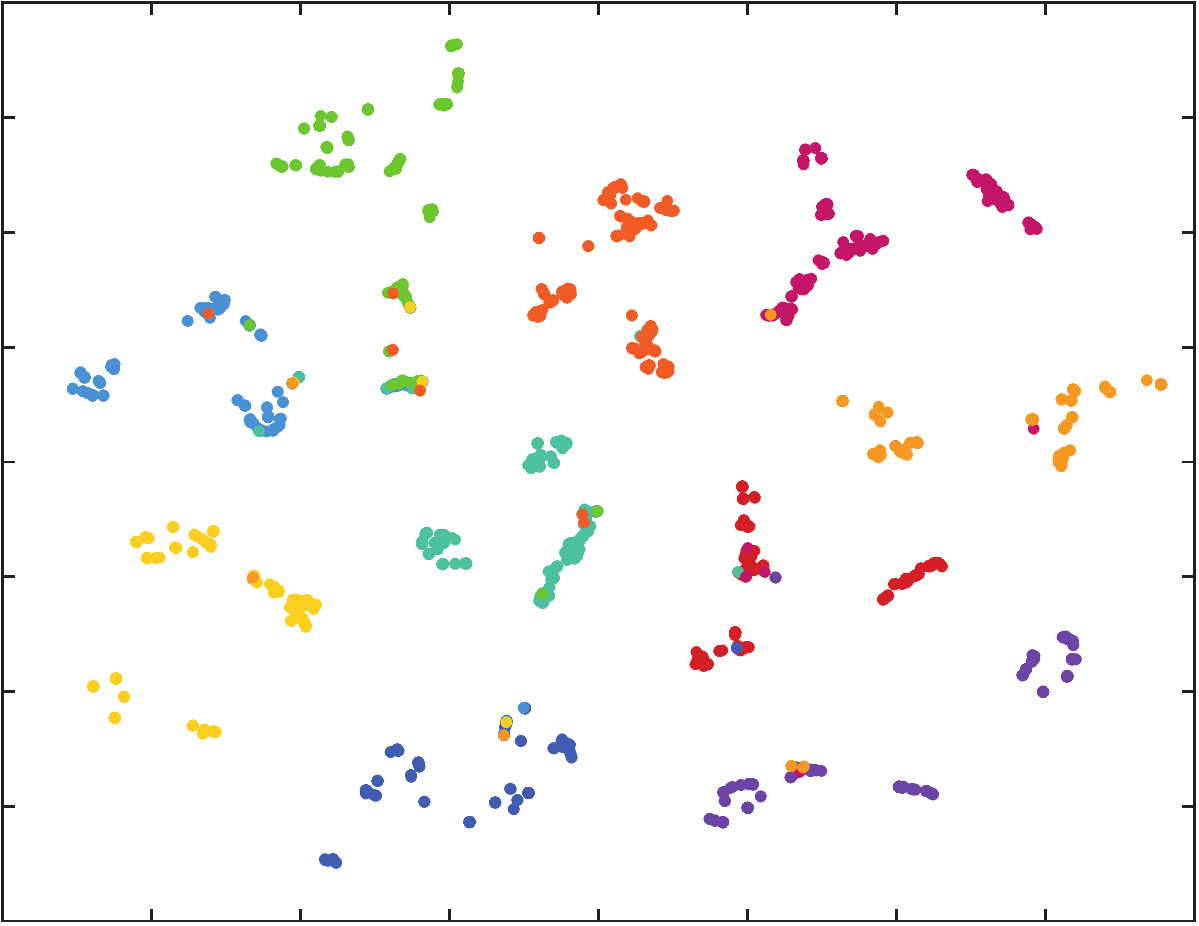}
			\caption{Ours}
		\end{subfigure}
		\vspace{-0.2cm}
		\caption{t-SNE visualization of 32 bits quantized features between DQN, DVSQ and DRQ. Images are randomly sampled on CIFAR-10 dataset, and samples with different labels are marked with different colors.}
		\label{fig:tSNE}
		\vspace{-0.2cm}
	\end{center}
\end{figure*}

\subsubsection{Results}
On CIFAR-10, we compare our DRQ with a few state-of-the-art methods, including DRSCH~\cite{DRSCH}, DSCH~\cite{DRSCH}, DSRH~\cite{DSRH}, VDSH~\cite{VDSH}, DPSH~\cite{DPSH}, DTSH~\cite{DPSH}, DTSH~\cite{DTSH}, DSDH~\cite{DSDH} and PQNet~\cite{PQNet}, using 16, 24, 36, 48 bits. We set $M=4$ and $K = 2^{\frac{L}{M}} = 16, 64, 512, 2048$. The results on CIFAR dataset are shown in Table~\ref{tab:eccv_cifar}. Results show our network achieves comparable mAP performance against state-of-the-art methods, \ie PQNet. Our mAP is only 0.3\%-0.5\% lower than PQNet. Results also show our performance is stable with variable bit-lengths. Our method only get 0.1\% decrease when bit-length shrinks to 16 bits. Noticed that our recurrent quantization only use a single codebook with $K$ codewords, and therefore our method requires much less parameters compared with other methods, as shown in Tab.~\ref{tab:SizeCompare}.

On NUS-WIDE dataset, we compare our method with a few shallow and deep methods. Shallow methods include SH~\cite{SH}, ITQ~\cite{ITQ}, LFH~\cite{LFH}, KSH~\cite{KSH}, SDH~\cite{SDH}, FASTH~\cite{FastH}\eat{, we evaluate their performance based on unsupervised deep neural network \ie we use a pretrained AlexNet to extract features and use shallow methods to compress features}. Deep methods include NINH~\cite{NINH}, DHN~\cite{DHN}, DQN~\cite{DQN}, DPSH~\cite{DPSH}, DTSH~\cite{DTSH}, DSDH~\cite{DSDH} and PQNet~\cite{PQNet}. The results are generated in 12, 24, 36, 48 bits. We fix $K=2048, M=4$ to generate 48 bits codes, and then slice these codes to get shorter binary codes. The results are shown in Table~\ref{tab:eccv_nus}. On NUS-WIDE, our method achieves the highest mAP compared with state-of-the-art methods when the code length is longer than 12 bits. Noticed that our method uses a shared codebook for all code lengths, and it is trained once.

The codebook size \wrt code-length comparison between multiple methods is shown in Tab.~\ref{tab:SizeCompare}. Our method obtains the smallest codebook size compared with the other methods. Also, to generate binary codes with different lengths, our methods is trained once.

\begin{table}[t]
	\centering
	\small
	\resizebox{1\columnwidth}{!}{%
		\begin{tabular}{c|cccccc}
			\hline
			Methods    & 8 bits  & 16 bits  & 24 bits  & 32 bits  & 40 bits  & 48 bits  \\ \hline
			PQ (PQNet) & 524k    & 524k     & 524k     & 524k     & 524k     & 524k     \\
			OPQ        & 4.72M   & 4.72M    & 4.72M    & 4.72M    & 4.72M    & 4.72M    \\
			AQ (DTQ)   & 524k    & 1.05M    & 1.57M    & 2.10M    & 2.62M    & 3.15M    \\ \hline
			DRQ        &                    \multicolumn{6}{c}{\textbf{524k}}                   \\ \hline
		\end{tabular}
	}
	\caption{Codebook size \wrt code-length comparison among multiple methods (we set $D=2048$ and $K=256$). PQNet uses similar codebook structure to PQ, and so does DTQ to AQ.}
	\label{tab:SizeCompare}
\end{table}

\subsection{Comparison Results Using Setting 2}
\label{sec:Setup2}
\subsubsection{Setting}
Following DTQ~\cite{DTQ}, on \textbf{CIFAR-10}, we combine the training and validation set together, and randomly select 500 images per class as the training set, 100 images per class as the query set. The remaining images are used as the database. On \textbf{NUS-WIDE}, we use the subset of 195,834 images from the 21 most frequent concepts. We randomly sample 5,000 images as the query set, and use the remaining images as the database. Furthermore, we randomly select 10,000 images from the database as the training set. On \textbf{ImageNet}, we follow~\cite{DVSQ} to randomly choose 100 classes. We use all the images of these classes in the training set as the database, and use all the images of these classes in the validation set as the queries. Furthermore, we randomly select 100 images for each class in the database for training. We compare our method with 11 classical hash or quantization methods, including 5 shallow methods: ITQ-CCA~\cite{ITQ}, BRE~\cite{BRE}, KSH~\cite{KSH}, SDH~\cite{SDH} and SQ~\cite{StackedQuantizer}, and 6 deep architecture: CNNH~\cite{CNNH}, DNNH~\cite{DNNH}, DHN~\cite{DHN}, DSH~\cite{DSH}, DVSQ~\cite{DVSQ}, DTQ~\cite{DTQ}.

\subsubsection{Results}
We use mAP@54000 on CIFAR-10 and mAP@5000 on NUS-WIDE and ImageNet. We use 8, 16, 24, 32-bits codes by setting $M=4, K=256$. We also use precision-recall curve and precision@R (returned results) curve to evaluate the retrieval quality. The results are shown in Table~\ref{tab:Result}, Fig.~\ref{fig:CurveOnFirst} and Fig.~\ref{fig:CurveOnSecond}.

It can be observed that: 1) Our DRQ significantly outperforms the other methods in CIFAR-10 and NUS-WIDE datasets. Specifically, it outperforms the best counterpart (DTQ) by 1.8\%, 3.5\%, 4.2\% and 4.5\% on CIFAR-10, and by 1.2\%, 1.7\% and 2.0\% on NUS-WIDE dataset. DRQ is outperformed by DTQ on NUS-WIDE for 8-bit codes. The possible reason is that in DRQ, codebooks are shared by different code lengths, and may lose some accuracy especially for short binary codes. On ImageNet, our method is outperformed by DTQ, which may be caused by the random selection. Also, our DRQ requires much less parameters than DTQ, and our model is only trained once. 2) With the increase of code length, the performance of most indexing methods is improved accordingly. For our DRQ, the mAP increased by 3.4\%, 7.3\% and 3.6\% for CIFAR-10, NUS-WIDE and ImageNet dataset respectively. This verifies our DRQ can generate sequential binary codes to gradually improve accuracy. 3) The performances of precision-recall curves for different methods are consistent with their performances of mAP. The precision curves represent the retrieval precision with respect to number of return results.

\subsection{Ablation Study}
\label{sec:Ablation}
In this subsection, we study the effect of each part in our architecture using the following settings. 1) Unsupervised quantization: we use raw $fc7$ output without additional $fc8, fc9$. 2) Remove $L_s$: we remove $fc9$ and $L_s$, and change $fc8$ to 2048-d and apply $L_t$ on it directly, to validate the role of $L_s$. 3) Remove $L_t$: we remove $fc9$ and $L_t$ and change $fc8$ to 300-d with $L_s$, to validate the role of $L_t$. 4) $\tilde{\textbf{q}}$ only: we remove the construction of $\hat{\textbf{q}}$ and  associated losses, to validate the role of $\hat{\textbf{q}}$. 5) Remove $E_j$: we remove $E_j$ to validate the effectiveness of joint central loss. 6) Intermediate supervision: we use only $fc9$, which is 300-d and also apply $L_t$ to $fc9$, to validate effectiveness of the concatenation of $fc8, fc9$. We perform ablation study on NUS-WIDE and show results in Tab.~\ref{tab:Ablation}. In general, DRQ performs the best, and `Remove $\ell_s$' ranks the second. By removing $\ell_s$, the supervision information is still utilized in $\ell_t$, so mAP drop is not that significant. The unsupervised architecture also achieves good results, indicating the validity of pretrained AlexNet. Notice that if we remove any of the objective functions in the structure, mAP will have a huge loss. This indicates the effectiveness of each part of our DRQ. We get the worst result when we remove the concat, this may be of the significant information loss in 300-d features.
\begin{table}[t]
	\centering
	\small
	\begin{tabular}{c|cccc}
		\hline
		Structure        & 8 bits & 16 bits & 24 bits & 32 bits \\ \hline
		Unsupervised     & 0.582 & 0.630  & 0.629  & 0.626  \\
		Remove $\ell_s$     & \textbf{0.750} & \underline{0.759}  & \underline{0.761}  & \underline{0.762}  \\
		Remove $\ell_t$     & 0.629 & 0.672  & 0.678  & 0.680  \\
		$\tilde{\textbf{q}}$ only & 0.654 & 0.725  & 0.731  & 0.734  \\
		Remove $E_j$     & 0.604 & 0.628  & 0.628  & 0.628  \\
		Remove concat    & 0.544 & 0.569  & 0.571  & 0.575  \\ \hline
		DRQ  & \underline{0.748} & \textbf{0.810}  & \textbf{0.817}  & \textbf{0.821} \\ \hline
	\end{tabular}
	\caption{Ablation study of our method on NUS-WIDE dataset.}
	\label{tab:Ablation}
\end{table}

\subsection{Qualitatively Results}

To qualitatively validate the performance of quantization methods, we also perform t-SNE visualization on DTQ, DVSQ and DRQ, and show the results in Fig.\ref{fig:tSNE}. Visualizations are created on CIFAR-10, we randomly sample 5,000 images from database and adopt 32 bits quantized features. Our DRQ has a similar performance to DTQ, and they both show distinct clusters in their visualization, which is much better than DVSQ. Our unsupervised structure also has a promising performance since data points are concretely clustered. However, some of the data points with different labels are wrongly clustered together. This indicates the importance of supervision information.

\section{Conclusion}
In this paper, we propose a Deep Recurrent Quantization (DRQ) architecture to generate sequential binary codes. When the model is trained once, a sequence of binary codes can be generated and the code length can be easily controlled by adjusting the number of recurrent iterations.
A shared codebook and a scalar factor is designed to be the learnable weights in the deep recurrent quantization block, and the whole framework can be trained in an end-to-end manner.
Experimental results on the benchmark datasets show that our model achieves comparable or even better performance compared with the state-of-the-art for image retrieval, but with much less parameters and training time.

\section*{Acknowledgements}
This work is supported by the Fundamental Research Funds for the Central Universities (Grant No.~ZYGX2014J063, No.~ZYGX2016J085), the National Natural Science Foundation of China (Grant No.~61772116, No.~61872064, No.~61632007, No.~61602049).
{
\bibliographystyle{named}
\bibliography{ijcai19}
}
\end{document}